\documentclass[a4paper,fleqn]{cas-dc}
\usepackage[english]{babel}
\usepackage{csquotes}
\usepackage[numbers,sort&compress]{natbib}
\usepackage{siunitx}
\usepackage{tabularx}
\usepackage{anyfontsize}
\usepackage{placeins}
\usepackage[nospread,noshrink]{flushend}
\usepackage{needspace}
\usepackage{xurl}
\usepackage{microtype}
\usepackage{CharisSIL}
\ifdefined\pdfmapfile\pdfmapfile{+charssil.map}\fi
\ifdefined\pdfminorversion\pdfminorversion=7\fi
\ifdefined\pdfinfoomitdate\pdfinfoomitdate=1\fi
\ifdefined\pdfsuppressptexinfo\pdfsuppressptexinfo=-1\fi
\ifdefined\pdftrailerid\pdftrailerid{}\fi
\definecolor{RamanDeep}{HTML}{155E6B}
\definecolor{RamanMid}{HTML}{2D8793}
\definecolor{RamanRow}{HTML}{E7F3F4}
\definecolor{RegressionBand}{HTML}{EAF3F8}
\definecolor{ClassificationBand}{HTML}{EEF5EA}
\definecolor{OverallBand}{HTML}{FFF2DE}
\definecolor{GroupBand}{HTML}{F3F4F5}
\definecolor{SecondTone}{HTML}{4B4B4B}

\newcolumntype{L}[1]{>{\raggedright\arraybackslash}p{#1}}
\newcolumntype{C}[1]{>{\centering\arraybackslash}p{#1}}

\newcommand{\bestnum}[1]{{\bfseries #1}}
\newcommand{\secondnum}[1]{\multicolumn{1}{c}{\bfseries\color{SecondTone}\underline{#1}}}
\newcommand{\componenton}{\textcolor{RamanDeep}{\checkmark}}
\newcommand{\componentoff}{\textcolor{black!25}{\ensuremath{\circ}}}
\newcommand{\groupheading}[2]{%
  \rowcolor{GroupBand}\multicolumn{#1}{@{}l}{\textit{#2}}\\
}

\usepackage{arxiv-layout}
\newcommand{\Cref}[1]{\autoref{#1}}

\providecommand{\doi}[1]{\url{https://doi.org/#1}}
\hypersetup{
  pdftitle={},
  pdfauthor={},
  pdfsubject={},
  pdfkeywords={},
  pdfcreator={},
  pdfproducer={},
  bookmarksnumbered=true,
  bookmarksopen=true
}
\begin{document}
\shorttitle{RamanPFN}
\shortauthors{X. Pan et al.}
\title[mode=title,size=\fontsize{13.5}{17.25}\selectfont]{RamanPFN: A physics-guided spectral foundation framework for in-context learning across diverse Raman applications}
\author[1,2]{Xingyu Pan}
\ead{panxingyu2003@buaa.edu.cn}
\author[2]{Huan Wang}
\cormark[1]
\ead{wh.2021@tsinghua.org.cn}
\author[2]{Jinjia Guo}
\ead{guojj.jackey2024@gdhfi.com}
\author[2]{Zhenlin Zhao}
\ead{zhenlinzhao@cleerlab.com}
\author[2]{Siming Dong}
\ead{william.d@cleerlab.com}
\author[1]{Jixi Lu}
\ead{lujixi@buaa.edu.cn}
\affiliation[1]{organization={Beihang University},city={Beijing},country={China}}
\affiliation[2]{organization={Cleer Science},city={Beijing},country={China}}
\cortext[1]{Corresponding author.}
\begin{abstract}
Raman spectroscopy enables label-free molecular characterization across materials science, analytical chemistry, biomedicine, and industrial process monitoring.
However, machine learning for high-dimensional spectroscopy remains constrained by limited labeled data and a mismatch between the physical organization of spectra and feature-agnostic models.
Tabular foundation models such as TabPFN predict in context without parameter updates, but on wide spectra feature subsampling preserves channel coverage while losing joint visibility: in the five widest tasks, every channel was covered, yet only 4.4\% of random channel pairs co-occurred in any forward pass.
We present RamanPFN, a general-purpose spectral foundation framework for unified in-context inference through physics-guided spectral learning, which encodes the additive mixing of constituent spectra and the locality of vibrational bands.
Global Compositional Unmixing decomposes distributed, multi-band mixture signatures into shared non-negative bases to couple covarying peaks across the full spectrum, while Local Vibrational Subspace Encoding extracts orthogonal modes within contiguous spectral neighborhoods to preserve peak morphology, intensity fluctuations, and peak shifts.
Posterior marginalization over admissible unmixings unifies both scales into in-context prediction without updating the pretrained model.
Across 74 public Raman datasets exceeding 330{,}000 spectra, RamanPFN outperformed 26 methods spanning chemometrics, spectral neural networks, deep tabular learners, and tabular foundation models on 129 regression targets, reducing backbone RMSE by 13.2\%.
We further demonstrate across TabPFN-3, TabICLv2, and Kumo-Tabular that this advantage generalizes to both regression and classification tasks, reducing regression error by 12.3\%--25.7\% and multiclass log loss by 42.4\%--63.5\% to enable data-efficient quantitative analysis and sample identification.
\end{abstract}


\maketitle
\hypersetup{pdftitle={},pdfauthor={},pdfsubject={},pdfkeywords={},pdfcreator={},pdfproducer={}}
\begin{figure*}[pos=t]
\input{figure_1_framework}
\end{figure*}
\section{Introduction}

Raman spectroscopy provides label-free access to molecular composition and structure across materials science, analytical chemistry, biomedicine, and industrial process monitoring.
Its applications range from nanoscale materials characterization and molecular imaging to biochemical quantification, disease assessment, and the monitoring of chemical and biological processes~\citep{yang2021ters,li2025fire,kamp2024fumarate,huang2023liver,zhang2024singlecell,ilchenko2024miniaturization}.
Across these settings, spectral analysis estimates component concentrations and material properties, distinguishes chemically similar samples, and detects changes in molecular or physiological states.
Advances in acquisition speed and instrument miniaturization are extending these measurements to larger sample collections and increasingly diverse experimental environments~\citep{li2025fire,ilchenko2024miniaturization}.
However, extracting predictive information remains challenging since molecular specificity is expressed through complex, overlapping spectral bands whose positions, shapes, and relative intensities vary with the local chemical environment~\citep{georgiev2024,fan2023deepraman,sigle2023spectromics}.
Instrumental variation, substrate effects, and sample heterogeneity further complicate the separation of target-relevant chemical signatures from other sources of spectral variation~\citep{hu2024rspssl,gong2024cfnet}.
Learning these relationships is particularly difficult when thousands of correlated wavenumber channels are paired with relatively few labeled samples.
Establishing reference labels typically requires independent analytical measurements or expert interpretation, adding substantial costs to supervised data collection~\citep{ilchenko2024miniaturization,huang2023liver}.
Under these constraints, reliable prediction hinges on how effectively learning methods capture the physical and chemical relationships encoded across the spectrum.

Methodological advances in spectral analysis have largely focused on how this physical structure is represented.
In classical chemometrics, partial least-squares (PLS) regression and discriminant analysis project high-dimensional channels onto supervised latent directions and remain competitive in small-sample regimes~\citep{wold2001pls,barker2003plsda,xie2022plssubsampling}.
Spectral unmixing introduces non-negativity and other physically motivated constraints to decompose full-spectrum variance into component profiles and abundance-like coordinates~\citep{lee1999nmf,georgiev2024,rasti2024hysupp,gu2024stretchednmf}.
While these compact global representations are data-efficient, they can under-resolve nonlinear peak distortions, broadening, and subtle chemical shifts.
Conversely, deep spectral networks provide the capacity to model complex nonlinear and local patterns for molecular identification, imaging, and disease diagnosis~\citep{horgan2021deeper,huang2023liver,fan2023deepraman,lin2021srs,wan2025ipsc,zhu2025openset,zhao2024plmaf}.
However, these deep architectures require task-specific training, where limited reference labels heighten the risk of learning instrumental artifacts instead of genuine chemical signatures.
Self-supervised learning has further advanced spectral preprocessing through denoising and baseline correction~\citep{hu2024rspssl,xue2025spectralfm}.
It has also enabled spectral feature learning from unlabeled data by reconstructing masked Raman spectra during pretraining and fine-tuning the network for classification with limited labeled data~\citep{ren2025smae}.
Even so, reliably extracting predictive spectral relationships under scarce supervision remains an open challenge.

Tabular foundation models offer a fundamentally different approach by acquiring the learning algorithm itself during pretraining~\citep{hollmann2025}.
By executing in-context prediction on unseen datasets without parameter updates, models in the TabPFN and TabICL families have achieved strong small-data performance across tabular domains~\citep{hollmann2025,tabicl2025,mcelresh2023tabzilla,tabarena2025} and have also demonstrated their effectiveness in Raman prediction~\citep{ramanbench}.
Nevertheless, their underlying architectures are designed for generic tabular data, leaving both spectral adjacency and full-spectrum compositional correlations implicit~\citep{reiter2026nir}.
To accommodate wide inputs exceeding native feature limits, current frameworks rely on feature subsampling, dimensionality reduction, or continued pretraining~\citep{tabdpt2025,ye2025closer,liu2025tabpfnunleashed,kolberg2025tabpfnwide,tunetables2024,gotabpfn2026}.
Feature subsampling can separate channels that define vibrational envelopes or distributed molecular signatures across different forward passes.
Coverage is not context.
In our analysis of the five widest tasks, every channel was covered across the forward passes, yet only 4.4\% of randomly sampled channel pairs appeared together in at least one forward pass.

Raman spectra encode chemical information through both full-spectrum compositional relationships and local vibrational structure.
At the global scale, a single molecular constituent contributes vibrational bands at widely separated wavenumbers, so changes in its abundance produce correlated responses across distant spectral regions.
At the local scale, variations in molecular conformation, chemical environment, and physical state are expressed through changes in peak intensity, shape, shoulder structure, and position within contiguous spectral regions.
Foundation models across scientific domains incorporate 3D atomic geometry in materials science~\citep{deng2023chgnet,merchant2023gnome,zeni2025mattergen}, temporal order in forecasting~\mbox{\citep{das2024timesfm,ansari2024chronos}}, and spectral and spatial relationships in remote sensing~\citep{hong2024spectralgpt,hypersigma2025}.
Physics-guided machine learning incorporates scientific relationships and constraints into the design of learning frameworks~\citep{karpatne2017theoryguided,willard2023scientific}.
For Raman spectroscopy, this motivates a unified dual-scale learning framework that couples long-range compositional covariation while preserving multiple modes of local spectral variation.

In this paper, we propose RamanPFN, a general-purpose spectral foundation framework that establishes physics-guided spectral learning for unified in-context inference, as outlined in \Cref{fig:framework}.
RamanPFN bridges the structural gap between ordered physical spectra and tabular foundation models by explicitly parameterizing this dual-scale organization.
To capture full-spectrum compositional covariation, Global Compositional Unmixing (GCU) decomposes whole-spectrum profiles into non-negative component bases under the additive-mixing principle and projects each spectrum onto these fixed bases to obtain global compositional coordinates.
In parallel, Local Vibrational Subspace Encoding (LVSE) partitions the ordered wavenumber axis into contiguous intervals and extracts multiple orthogonal modes within each region, preserving localized variations in band intensity and shape, shoulder structure, and peak position.
The two sets of coordinates form a compact joint input that can be observed in full within each forward pass of a pretrained tabular foundation model, whose parameters remain fixed.
Since non-negative unmixing admits multiple solutions, RamanPFN marginalizes its predictions over several admissible unmixings.

We evaluated RamanPFN across 74 diverse public Raman datasets comprising more than 330{,}000 spectra from four application domains.
In summary, our main contributions are as follows:
\begin{enumerate}[(1)]
\item To the best of our knowledge, we provide the first systematic quantification of the spectral context gap in tabular foundation models that rely on feature subsampling. We show that joint channel visibility collapses with increasing spectral dimensionality despite complete channel coverage, revealing a structural limitation of feature-subsampled inference on high-dimensional spectra.
\item We propose RamanPFN, a physics-guided spectral foundation framework that integrates GCU and LVSE into in-context learning. By jointly capturing distributed molecular signatures and fine-grained peak variations, the framework enables tabular foundation models to exploit the compositional and vibrational structure encoded across the spectrum.
\item We demonstrate RamanPFN's effectiveness across 129 regression targets against 26 independently reproduced methods spanning chemometrics, spectral neural networks, deep tabular learners, and tabular foundation models. RamanPFN achieves the lowest aggregate root-mean-square error (RMSE) and mean absolute error (MAE), ranks first in Score and Elo, and reduces error by 13.2\% relative to its pretrained backbone, with dataset-level gains over all 26 baselines significant after Holm correction.
\item We further show that this advantage transfers across tabular foundation models for both quantitative analysis and sample identification. Applied with the same configuration to TabPFN-3, TabICLv2, and Kumo-Tabular across 114 regression targets and 21 classification tasks, RamanPFN reduces regression error by 12.3\%--25.7\% and multiclass log loss by 42.4\%--63.5\% relative to each native model.
\end{enumerate}
\flushcolsend
\wideblock{\begin{widefigure}
    \centering
    \includegraphics[width=\textwidth]{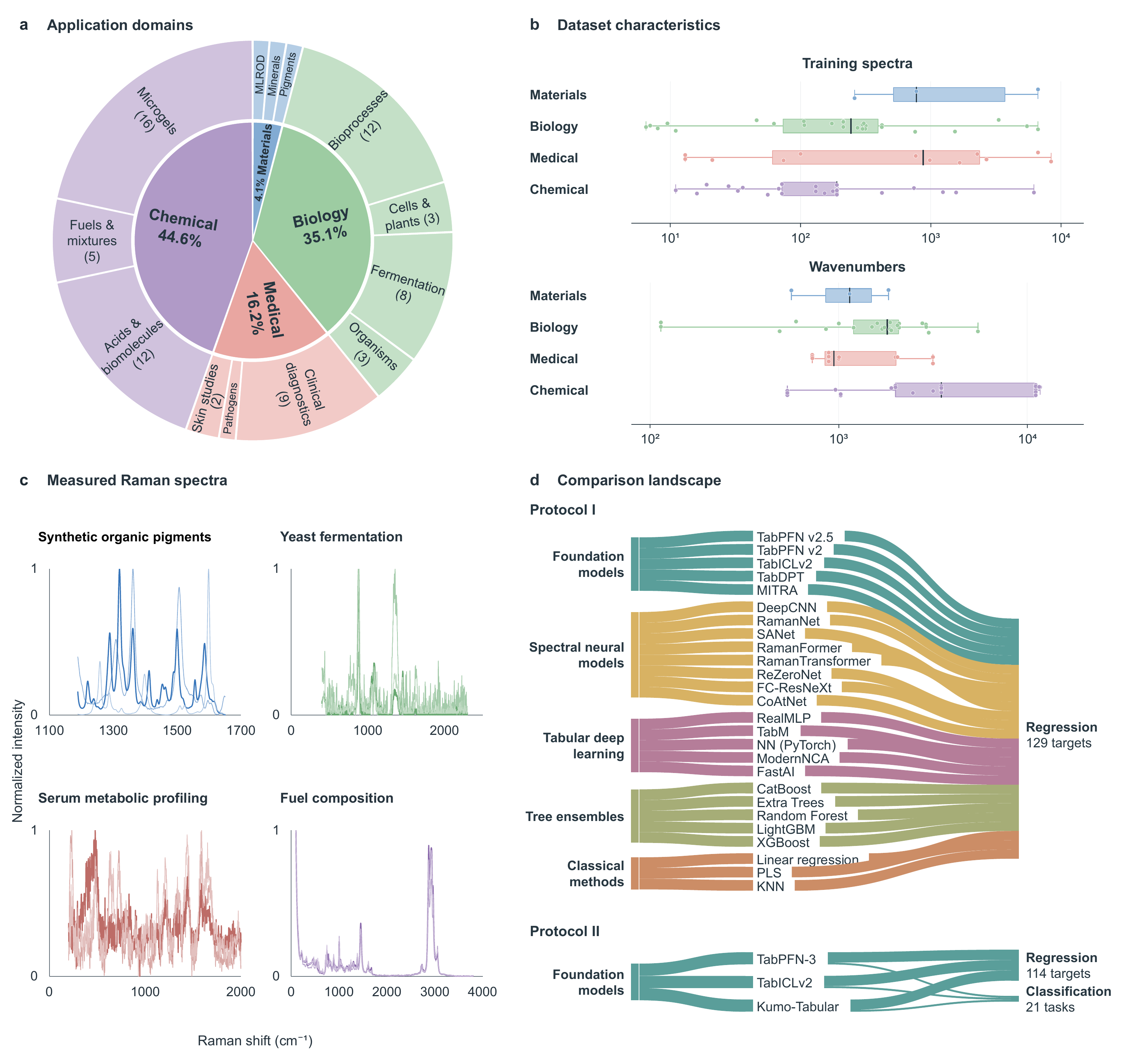}
    \caption{Dataset collection and dual-protocol evaluation design.
    (a)~Distribution of the 74 public Raman datasets across four application domains, with dataset counts shown for multi-dataset groups.
    (b)~Training-set sizes and wavenumber channel counts per dataset on logarithmic axes.
    (c)~Representative min--max-normalized spectra from four datasets.
    (d)~Comparison settings of Protocol I across 26 baselines and Protocol II across three tabular foundation models.}
    \label{fig:evaluation_landscape}
\end{widefigure}
}

\Needspace{7\baselineskip}
\section{RamanPFN}
\label{sec:ramanpfn}

RamanPFN establishes physics-guided spectral learning by modeling two fundamental properties of Raman spectroscopy: mixture spectra arise from approximately additive, non-negative combinations of constituent responses across the full spectrum, whereas subtle vibrational perturbations manifest within contiguous wavenumber neighborhoods. To capture both scales, RamanPFN pairs Global Compositional Unmixing (GCU) with Local Vibrational Subspace Encoding (LVSE). As outlined in \Cref{fig:framework}, the resulting global and local coordinates are concatenated into a compact joint input for in-context inference by a pretrained tabular foundation model, whose predictions are then marginalized over admissible unmixings.

\wideblock{\begin{widetable}
\centering
\rmfamily
\caption{Raman dataset collection and task characteristics.}
\label{tab:dataset_collection}
\footnotesize
\setlength{\tabcolsep}{3.0pt}
\renewcommand{\arraystretch}{1.17}

\textbf{(a)}\quad\textbf{Scientific coverage by application domain}\par\vspace{2pt}
\begin{tabularx}{\textwidth}{@{}L{2.55cm} X
  S[table-format=2.0]
  S[table-format=3.0]
  S[table-format=3.0]
  S[table-format=2.0]@{}}
\toprule
\multirow{2}{*}{\textbf{Application domain}} &
\multirow{2}{*}{\textbf{Representative coverage}} &
{\multirow{2}{*}{\textbf{Datasets}}} &
{\textbf{Protocol I}} &
\multicolumn{2}{c}{\textbf{Protocol II}} \\
\cmidrule(lr){4-4}\cmidrule(l){5-6}
& & &
{\shortstack{\textbf{Regression}\\\textbf{targets}}} &
{\shortstack{\textbf{Regression}\\\textbf{targets}}} &
{\shortstack{\textbf{Classification}\\\textbf{tasks}}} \\
\midrule
Materials science & MLROD, RRUFF minerals, and synthetic organic pigments & 3 & 1 & 0 & 3 \\
Biological \& biotechnological & Bioprocess monitoring, bacterial and yeast fermentations, cancer-cell metabolites, locust hemolymph, marine pathogens, and mutant wheat & 26 & 60 & 60 & 6 \\
Medical \& clinical & Serum and saliva diagnostics, pathogenic bacteria, skin measurements, and blood glucose & 12 & 0 & 1 & 11 \\
Chemical \& industrial & Gasoline properties, organic acids, microgels, sugar mixtures, adenine, amino acids, and chlorinated samples & 33 & 68 & 53 & 1 \\
\midrule
\textbf{Total} & \textbf{74 public Raman datasets across four application domains} & 74 & 129 & 114 & 21 \\
\bottomrule
\end{tabularx}

\vspace{7pt}
{\fontsize{7.2}{9.6}\selectfont
\setlength{\tabcolsep}{1.7pt}
\textbf{(b)}\quad\textbf{Task and spectral characteristics}\par\vspace{2pt}
\begin{tabular*}{\textwidth}{@{\extracolsep{\fill}}l c c c c c c c c c c@{}}
\toprule
\multirow{2}{*}{\textbf{Task family}} &
\multirow{2}{*}{\textbf{Datasets}} &
\multirow{2}{*}{\textbf{Tasks}} &
\multicolumn{2}{c}{\textbf{Training spectra}} &
\multicolumn{2}{c}{\textbf{Test spectra}} &
\multicolumn{2}{c}{\textbf{Wavenumbers}} &
\multicolumn{2}{c}{\textbf{Output space}} \\
\cmidrule(lr){4-5}\cmidrule(lr){6-7}\cmidrule(lr){8-9}\cmidrule(lr){10-11}
& & &
\textbf{Range} & \textbf{Median} &
\textbf{Range} & \textbf{Median} &
\textbf{Range} & \textbf{Median} &
\textbf{Range} & \textbf{Median} \\
\midrule
Protocol I: regression & 53 & 129 & 6--6,203 & 143 & 2--1,668 & 36 & 114--11,689 & 1,901 & \multicolumn{2}{c}{Continuous} \\
Protocol II: regression & 40 & 114 & 5--5,430 & 110 & 2--2,827 & 55 & 114--11,689 & 1,901 & \multicolumn{2}{c}{Continuous} \\
Protocol II: classification & 21 & 21 & 13--8,401 & 767 & 6--4,201 & 384 & 561--3,160 & 1,142 & 2--79 classes & 3 classes \\
\bottomrule
\end{tabular*}
}
\par\vspace{4pt}
\begin{minipage}{\textwidth}
\footnotesize\raggedright
Note: (a)~Datasets and evaluated tasks by application domain across both protocols. (b)~Ranges and medians over the data splits of Protocol I and the outer folds of Protocol II.
\end{minipage}
\end{widetable}
}
\subsection{Problem formulation}

Let $X_{\mathrm{tr}}\in\mathbb{R}^{n\times D}$ and $X_{\mathrm{te}}\in\mathbb{R}^{m\times D}$ denote $n$ training and $m$ query spectra over $D$ ordered channels $\nu_1<\cdots<\nu_D$, with training targets $y_{\mathrm{tr}}$. Channel positions correspond to physical Raman shifts where available and to the recorded channel order otherwise. A tabular foundation model treats each spectrum $\mathbf{x}=[I(\nu_1),\ldots,I(\nu_D)]$ as a row of independent features. When $D$ exceeds its native feature limit, the model evaluates subsampled views $\mathbf{x}^{(j)}=[I(\nu_i)]_{i\in\mathcal{S}_j}$, $\mathcal{S}_j\subseteq\{1,\ldots,D\}$, in separate forward passes, as illustrated in \Cref{fig:framework}a. The views jointly cover all channels, yet each forward pass observes only a fragment of the spectrum, which separates distant covarying bands and breaks the continuity of local bands.

RamanPFN instead maps the spectra into global compositional coordinates $G$ and local vibrational coordinates $L$ that preserve both physical scales in a compact form. All decomposition bases and normalization statistics are estimated from the training spectra and applied unchanged to the query spectra.

\subsection{Global compositional unmixing}

Chemical constituents in a mixture typically give rise to multiple Raman bands whose intensities covary across the full spectrum. GCU captures this distributed structure through an additive, non-negative decomposition of the spectral matrix. A non-negative mapping $\varphi$, consisting of deterministic shifts and scaling, first transforms the spectra into $X_b^{+}=\varphi(X_b)$, $b\in\{\mathrm{tr},\mathrm{te}\}$. Global and channel-wise statistics of $\varphi$ are estimated from the training spectra, while spectrum-wise operations are applied independently to each sample.

Given an unmixing specification $\mathcal{U}=(\varphi,\rho)$ with non-negative mapping $\varphi$ and decomposition rank $\rho$, the transformed training spectra are decomposed by minimizing
\begin{equation}
\label{eq:02-1}
\min_{W_{\mathrm{tr}}\geq 0,\,H\geq 0}
\left\|X_{\mathrm{tr}}^{+}-W_{\mathrm{tr}}H\right\|_{F}^{2},
\end{equation}
where $W_{\mathrm{tr}}\in\mathbb{R}_{+}^{n\times\rho}$ contains sample coordinates and $H\in\mathbb{R}_{+}^{\rho\times D}$ contains spectral components. Each row of $H$ spans all $D$ channels, so distant covarying bands, such as bands A and B in \Cref{fig:framework}b, can share a single compositional coordinate.

The factors are initialized by non-negative double singular value decomposition (NNDSVD) and updated by hierarchical alternating least squares (HALS). The learned basis $H$ is then fixed, and every spectrum is projected onto it by non-negative least squares,
\begin{equation}
\label{eq:02-2}
G_{b}
=\underset{W\geq 0}{\arg\min}
\left\|X_{b}^{+}-WH\right\|_{F}^{2},
\qquad b\in\{\mathrm{tr},\mathrm{te}\},
\end{equation}
so that training and query spectra are expressed by abundance-like coordinates on a shared compositional basis.

\subsection{Local vibrational subspace encoding}

While GCU models full-spectrum compositional covariation, chemical shifts, hydrogen bonding, and concentration changes also perturb the fine morphology of individual Raman bands. LVSE resolves these localized variations within contiguous wavenumber neighborhoods. Each spectrum is first normalized by the standard normal variate (SNV) to remove spectrum-wise offset and scale, and then centered with the channel-wise training mean $\overline{\mathbf{x}}_{\mathrm{tr}}$:
\begin{equation}
\label{eq:02-3}
\widetilde X_{b}
=\mathrm{SNV}(X_{b})-\mathbf{1}\,\overline{\mathbf{x}}_{\mathrm{tr}}^{\mathsf T},
\qquad b\in\{\mathrm{tr},\mathrm{te}\}.
\end{equation}

The ordered channel axis is partitioned into contiguous intervals $\Omega_1,\ldots,\Omega_M$. Within interval $\Omega_m$, let $X_m^{b}=\widetilde X_{b}[:,\Omega_m]$ denote the local submatrix. A singular value decomposition of the training submatrix defines an orthogonal local basis,
\begin{equation}
\label{eq:02-4}
X_m^{\mathrm{tr}}=U_m\Sigma_mV_m^{\mathsf T},
\qquad
L_m^{b}=X_m^{b}\,V_{m,1:k_m},
\qquad b\in\{\mathrm{tr},\mathrm{te}\},
\end{equation}
where $V_{m,1:k_m}$ retains the $k_m$ leading right singular vectors. Multiple orthogonal modes per interval preserve local variations in band intensity, envelope shape, shoulder structure, and peak position that a single dominant direction would omit.

Concatenating the interval coordinates in spectral order yields the local vibrational coordinates
\begin{equation}
\label{eq:02-5}
L_{b}
=\left[L_1^{b}\,\middle\Vert\,L_2^{b}\,\middle\Vert\,\cdots\,\middle\Vert\,L_M^{b}\right],
\qquad b\in\{\mathrm{tr},\mathrm{te}\},
\end{equation}
where $\Vert$ denotes column-wise concatenation and each coordinate remains anchored to its spectral neighborhood $\Omega_m$.

\subsection{In-context inference with posterior marginalization}

Because neither the underlying number of chemical constituents nor the exact intensity scale of additive mixing is uniquely identifiable from unlabeled spectra alone, multiple admissible unmixing specifications $\mathcal{U}_1,\ldots,\mathcal{U}_R$ yield distinct yet physically plausible compositional coordinates $G_1,\ldots,G_R$. As shown in \Cref{fig:framework}c, the local and global coordinates of each unmixing are concatenated into a joint input,
\begin{equation}
\label{eq:02-6}
Z_{r}^{b}
=\left[L_{b}\,\middle\Vert\,G_{r}^{b}\right],
\qquad b\in\{\mathrm{tr},\mathrm{te}\},
\end{equation}
which is compact enough for a pretrained tabular foundation model $f_{\theta}$ to observe all local intervals and global components together in each forward pass. The model predicts the query targets in context,
\begin{equation}
\label{eq:02-7}
\hat{y}_{r}
=f_{\theta}\!\left(Z_{r}^{\mathrm{tr}},y_{\mathrm{tr}};Z_{r}^{\mathrm{te}}\right),
\end{equation}
where $\theta$ denotes the fixed pretrained parameters and $\hat{y}_{r}$ is a vector of predicted targets for regression or a matrix of class probabilities for classification.

Treating $\hat{y}_{r}$ as the posterior predictive estimate conditioned on $\mathcal{U}_r$, RamanPFN marginalizes over the admissible unmixings under a uniform prior, yielding the final prediction
\begin{equation}
\label{eq:02-8}
\hat{y}=\frac{1}{R}\sum_{r=1}^{R}\hat{y}_{r}.
\end{equation}

\section{Experiments}
\label{sec:experimental_setup}

\wideblock{\begin{widefigure}
    \centering
    \includegraphics[width=\textwidth]{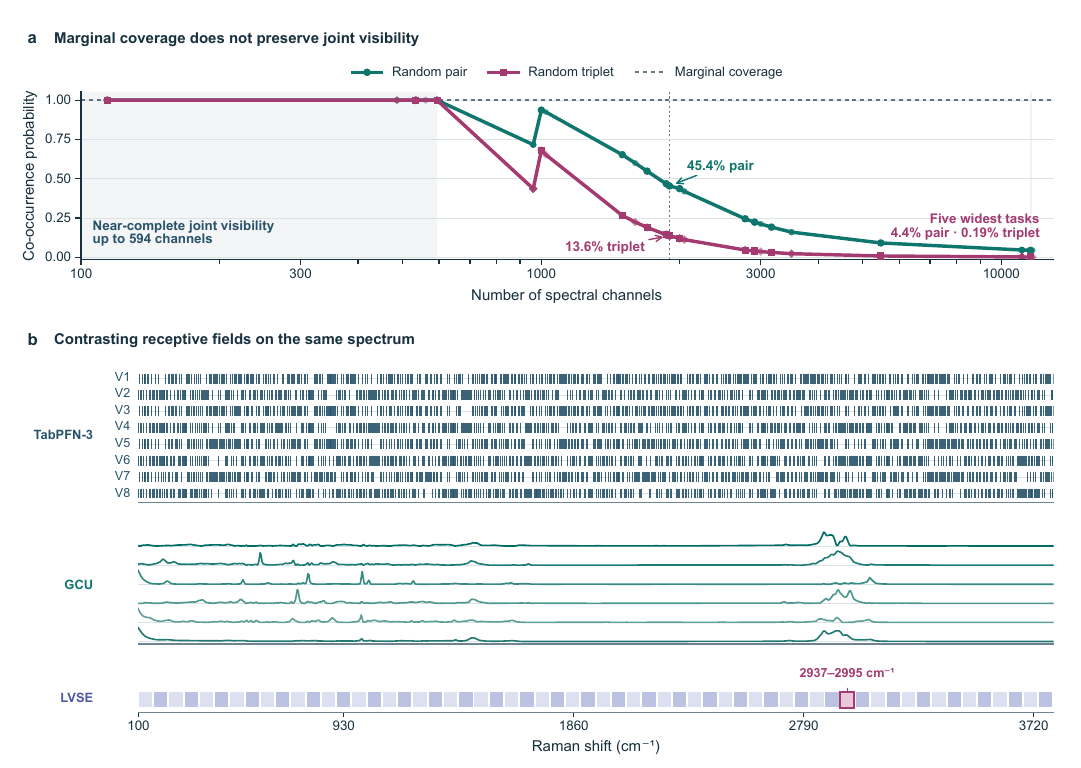}
    \caption{Channel co-occurrence and spectral receptive fields.
    (a)~Co-occurrence probabilities of random channel pairs and triplets across 129 regression tasks under TabPFN-3 feature subsampling, with the dotted vertical line marking the median width of 1{,}901 channels.
    (b)~Support maps on a 961-channel gasoline spectrum for eight TabPFN-3 views, six high-variance GCU components, and 61 contiguous LVSE intervals.}
    \label{fig:spectral_mechanism}
\end{widefigure}
}

\subsection{Datasets and tasks}

We evaluated RamanPFN on the 74 public RamanBench datasets, comprising more than 330{,}000 spectra~\citep{ramanbench}. As summarized in \Cref{tab:dataset_collection}a and mapped in \Cref{fig:evaluation_landscape}a, the collection spans four application domains, with 3 datasets in materials science, 26 in biological and biotechnological research, 12 in medical and clinical diagnostics, and 33 in chemical and industrial analysis. Each regression property or classification label was treated as a separate supervised task according to the published task definitions.

The task distribution in \Cref{fig:evaluation_landscape}b captures substantial variation in sample size and spectral dimensionality, while \Cref{fig:evaluation_landscape}c illustrates the corresponding diversity of measured spectral profiles. In Protocol I, the 129 regression targets from 53 datasets ranged from 6 to 6{,}203 training spectra and from 114 to 11{,}689 wavenumber channels, with a median task pairing 143 training spectra with 1{,}901 channels. Protocol II comprises 114 regression targets from 40 datasets and 21 classification tasks spanning 2 to 79 classes, of which 13 are multiclass and 8 are binary. \Cref{tab:dataset_collection}b summarizes the corresponding sample sizes and spectral dimensions.

\subsection{Evaluation protocols}

Protocol I used the regression tasks of the RamanBench v0.1 datasets with the published 80/20 train--test splits, with seeds 0, 1, and 2 defining three repetitions. Datasets with experimental groups were split with group-aware sampling, and the remaining datasets used seeded random sampling. Protocol II used the RamanBench v1.4 datasets to evaluate whether RamanPFN transfers across foundation-model architectures and prediction tasks, with each task evaluated over the three outer folds released with the datasets and group-aware splitting where experimental groups exist. Within each protocol, all methods were evaluated on identical task definitions and data partitions. Experiments were run with Python 3.12 and PyTorch 2.8 on NVIDIA GeForce RTX 4080 SUPER GPUs.

\wideblock{\begin{widefigure}
    \centering
    \includegraphics[width=\textwidth]{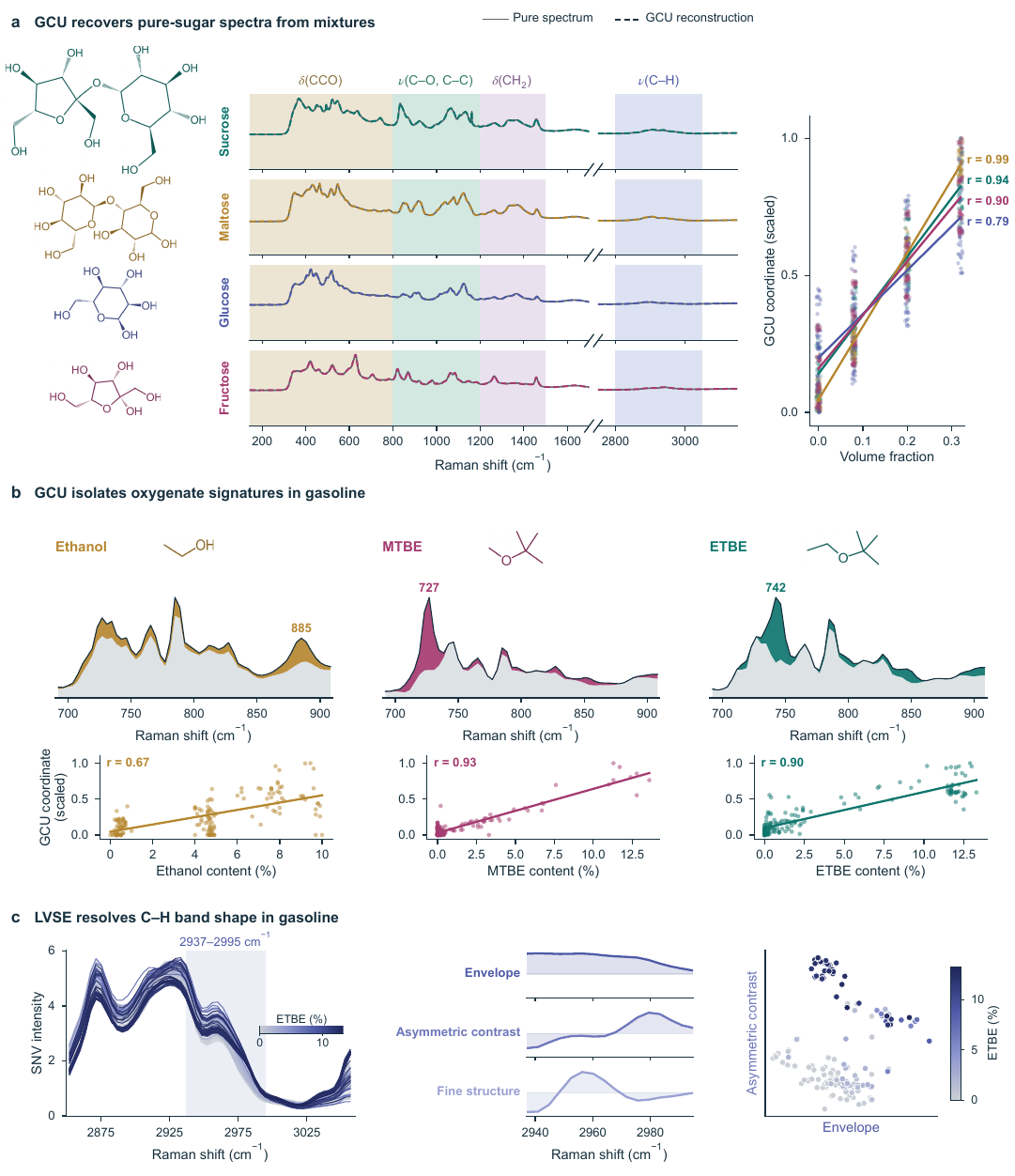}
    \caption{Physical interpretability of GCU and LVSE coordinates.
    (a)~Pure-sugar reference spectra, GCU reconstructions across shaded carbohydrate band regions, and relationships between scaled GCU coordinates and sugar volume fractions.
    (b)~Unsupervised GCU components associated with ethanol at 885~cm$^{-1}$, MTBE at 727~cm$^{-1}$, and ETBE at 742~cm$^{-1}$ in gasoline, with coordinate--content scatter plots below.
    (c)~Gasoline C--H stretching spectra colored by ETBE content, three orthogonal LVSE modes in 2{,}937--2{,}995~cm$^{-1}$, and the resulting local coordinate plane.}
    \label{fig:physical_interpretation}
\end{widefigure}
}
\subsection{Comparison methods, backbones, and ablation}

Under Protocol I, we independently reproduced 26 regression methods from the five families shown in \Cref{fig:evaluation_landscape}d, namely tabular foundation models, Raman-specific neural models, deep tabular and neural models, tree ensembles and gradient boosting, and classical machine learning and chemometrics. Each baseline retained its own preprocessing and fitting procedure on the common data partitions. For each target, the best baseline was the comparison method with the lowest RMSE after averaging the three repetitions.

RamanPFN and its ablation variants under Protocol I shared the pretrained TabPFN-3 backbone~\citep{tabpfn3_2026}. We evaluated four configurations across the 129 regression targets. The frozen backbone applied to normalized spectra served as the reference. The other three configurations augmented the backbone with GCU coordinates, used LVSE coordinates alone, and combined LVSE and GCU coordinates with marginalization over admissible unmixings in complete RamanPFN.

Under Protocol II, RamanPFN was applied with the same configuration to three pretrained tabular foundation models from distinct model families, TabPFN-3, TabICLv2~\citep{tabiclv2_2026}, and Kumo-Tabular. Each backbone with RamanPFN was compared with its own native inference on raw spectra across the regression, multiclass, and binary tasks.

\wideblock{\begin{widetable}
\centering
\rmfamily
\caption{Regression performance of RamanPFN and 26 comparison methods under Protocol I.}
\label{tab:regression_comparison}
\footnotesize
\setlength{\tabcolsep}{2.0pt}
\renewcommand{\arraystretch}{1.01}
\begin{tabular*}{\textwidth}{@{\extracolsep{\fill}}l
  S[table-format=2.3]
  S[table-format=2.3]
  S[table-format=2.1]
  S[table-format=2.2]
  S[table-format=1.4]
  S[table-format=4.0]@{}}
\toprule
\multirow{2}{*}{\textbf{Method}} &
\multicolumn{3}{c}{\cellcolor{RegressionBand}\textbf{Predictive error}} &
\multicolumn{3}{c}{\cellcolor{OverallBand}\textbf{Aggregate comparison}} \\
\cmidrule(lr){2-4}\cmidrule(lr){5-7}
& {\textbf{RMSE}$\downarrow$} & {\textbf{MAE}$\downarrow$} &
{\textbf{Gap to best (\%)}$\downarrow$} &
{\shortstack{\textbf{Average}\\\textbf{rank}$\downarrow$}} &
{\textbf{Score}$\uparrow$} & {\textbf{Elo}$\uparrow$} \\
\midrule
\groupheading{7}{Physics-guided spectral learning}
\rowcolor{RamanRow}[0pt][0pt]
\textbf{RamanPFN} & \bestnum{9.171} & \bestnum{6.910} & \bestnum{7.2} & \bestnum{2.65} & \bestnum{0.8694} & \bestnum{1,702} \\
\midrule
\groupheading{7}{Tabular foundation models}
TabPFN v2 & \secondnum{9.574} & 7.268 & 31.3 & \secondnum{5.29} & 0.6237 & \secondnum{1,489} \\
TabPFN v2.5 & 9.621 & \secondnum{7.227} & \secondnum{28.2} & 5.30 & \secondnum{0.6316} & 1486 \\
TabICLv2 & 10.141 & 7.642 & 34.8 & 7.12 & 0.5133 & 1386 \\
MITRA & 11.049 & 8.628 & 43.4 & 11.26 & 0.3357 & 1185 \\
TabDPT & 10.997 & 8.388 & 45.3 & 12.51 & 0.2453 & 1123 \\
\midrule
\groupheading{7}{Raman-specific neural models}
ReZeroNet & 11.224 & 8.675 & 42.9 & 11.19 & 0.2924 & 1178 \\
DeepCNN & 11.307 & 8.703 & 44.9 & 12.94 & 0.2376 & 1100 \\
RamanNet & 10.869 & 8.280 & 47.2 & 13.09 & 0.1934 & 1098 \\
RamanFormer & 10.964 & 8.430 & 48.9 & 14.54 & 0.2343 & 1025 \\
SANet & 12.625 & 10.063 & 56.4 & 18.17 & 0.0972 & 899 \\
RamanTransformer & 18.204 & 15.952 & 69.5 & 24.36 & 0.0063 & 565 \\
\midrule
\groupheading{7}{Deep tabular and neural models}
RealMLP & 11.198 & 8.465 & 46.0 & 12.92 & 0.2211 & 1148 \\
NN (PyTorch) & 10.750 & 8.221 & 47.9 & 13.59 & 0.1669 & 1105 \\
TabM & 10.812 & 8.270 & 50.5 & 14.22 & 0.1218 & 1089 \\
FC-ResNeXt & 11.578 & 8.804 & 49.4 & 14.95 & 0.1725 & 1048 \\
CoAtNet & 11.543 & 8.819 & 49.3 & 14.67 & 0.1554 & 1047 \\
FastAI & 12.573 & 9.828 & 59.4 & 20.84 & 0.0146 & 786 \\
ModernNCA & 13.308 & 10.121 & 60.6 & 21.60 & 0.0222 & 766 \\
\midrule
\groupheading{7}{Tree ensembles and gradient boosting}
CatBoost & 10.766 & 8.193 & 51.3 & 14.45 & 0.0888 & 1080 \\
Extra Trees & 10.330 & 7.873 & 50.7 & 14.29 & 0.1541 & 1071 \\
Random Forest & 10.517 & 8.008 & 52.7 & 16.04 & 0.1215 & 1000 \\
LightGBM & 11.385 & 8.781 & 55.4 & 18.13 & 0.0495 & 911 \\
XGBoost & 11.083 & 8.372 & 55.9 & 18.50 & 0.0743 & 899 \\
\midrule
\groupheading{7}{Classical machine learning and chemometrics}
Linear regression & 18.047 & 13.536 & 52.1 & 14.50 & 0.2124 & 1083 \\
PLS & 13.242 & 10.139 & 51.0 & 15.09 & 0.1813 & 1062 \\
KNN & 10.951 & 8.032 & 51.7 & 15.78 & 0.1549 & 1022 \\
\bottomrule
\end{tabular*}
\par\vspace{4pt}
\begin{minipage}{\textwidth}
\footnotesize\raggedright
Note: RMSE, root-mean-square error. MAE, mean absolute error. RMSE, MAE, and gap to best are means over 129 targets after averaging three repetitions, with gap to best defined as the mean target-wise value of $100[1-\mathrm{RMSE}_{\mathrm{best}}/\mathrm{RMSE}_m]$ for method $m$. Score and Elo are defined in \Cref{sec:evaluation_metrics}. Arrows indicate the direction of better performance. Bold values mark the best result in each column, and bold underlined values the second best.
\end{minipage}
\end{widetable}
}
\subsection{Implementation details}

A single fixed configuration was applied across all datasets, protocols, and backbones without task-specific tuning. In LVSE, the $D$ ordered channels were partitioned into contiguous equal-index intervals of width $\lceil D/64\rceil$, yielding at most 64 intervals, and at most three leading right singular vectors were retained per interval.

In GCU, two deterministic non-negative mappings were paired with two factorization ranks to define $R=4$ admissible unmixings. The first mapping subtracted the training channel-wise minimum, normalized each spectrum by its integrated area, and applied global scaling, whereas the second applied a global non-negative shift followed by global scaling. The nominal ranks were 16 and 32, each capped at the largest value in $\{2,3,4,6,8,12,16,24,32\}$ not exceeding $\min(n-1,D)$. Each basis was initialized by NNDSVDa and updated for 200 HALS iterations on the training spectra, and training and query spectra were projected onto the fixed basis by non-negative least squares with at most 400 iterations and a relative tolerance of $10^{-9}$.

Each of the four unmixings was concatenated with the shared LVSE coordinates as in Eq.~\eqref{eq:02-6}. We used two estimators per unmixing for TabPFN-3 and TabICLv2, giving eight estimators in total, and four per unmixing for Kumo-Tabular, giving sixteen. Predictions were averaged equally across the four unmixings, with predicted targets averaged for regression and class probabilities for classification, and all pretrained parameters were kept fixed.

\wideblock{\begin{widefigure}
    \centering
    \includegraphics[width=\textwidth]{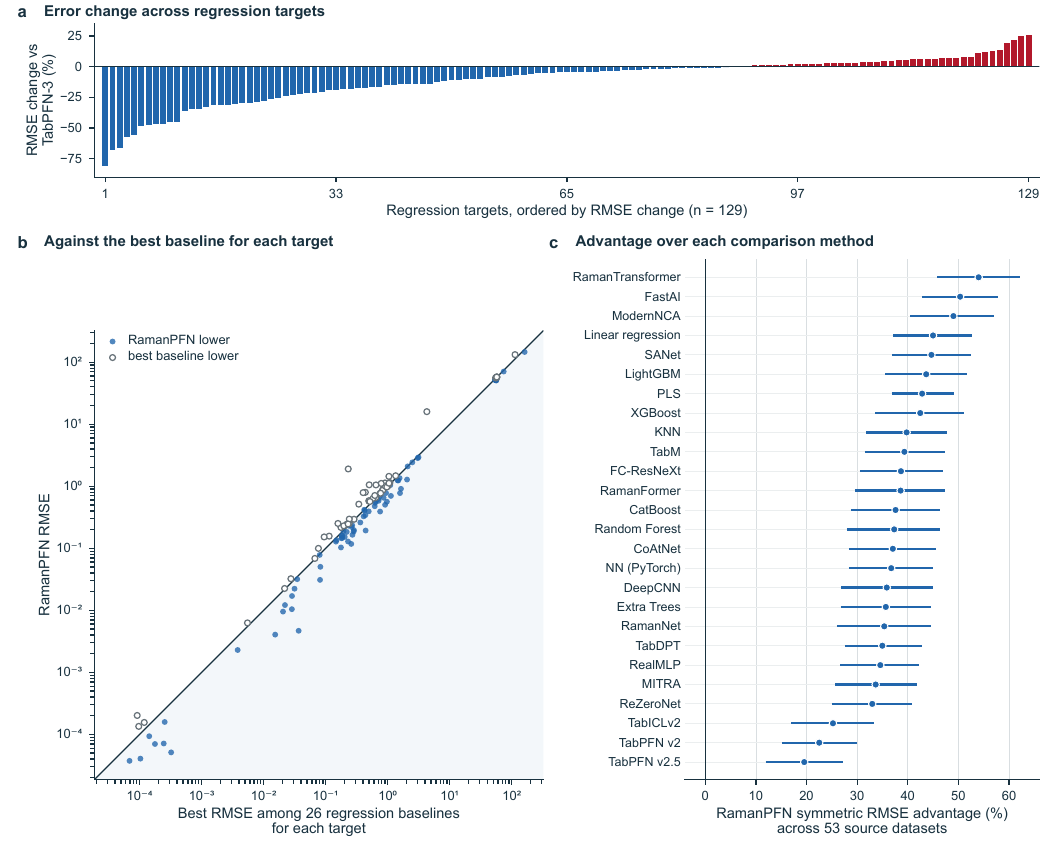}
    \caption{Regression performance of RamanPFN under Protocol I.
    (a)~Target-level RMSE changes relative to the frozen TabPFN-3 backbone across 129 regression targets, averaged over three repetitions.
    (b)~Repetition-averaged RMSE of RamanPFN against the strongest of 26 baselines on each target.
    (c)~Mean symmetric RMSE advantage over each baseline across 53 source datasets, with horizontal bars denoting 95\% bootstrap confidence intervals.}
    \label{fig:regression_performance}
\end{widefigure}
\par\vspace{10pt}\begin{widetable}
\centering
\rmfamily
\caption{Ablation analysis of GCU and LVSE under Protocol I.}
\label{tab:regression_ablation}
\footnotesize
\setlength{\tabcolsep}{2.0pt}
\renewcommand{\arraystretch}{1.20}
\begin{tabular*}{\textwidth}{@{\extracolsep{\fill}}l c c
  S[table-format=1.3]
  S[table-format=2.1]
  c c@{}}
\toprule
\multirow{2}{*}{\textbf{Configuration}} &
\multicolumn{2}{c}{\cellcolor{GroupBand}\textbf{Active mechanism}} &
\multicolumn{4}{c}{\cellcolor{RegressionBand}\textbf{Regression performance}} \\
\cmidrule(lr){2-3}\cmidrule(lr){4-7}
& {\textbf{GCU}} & {\textbf{LVSE}} &
{\shortstack{\textbf{Geometric mean}\\\textbf{RMSE ratio}$\downarrow$}} &
{\shortstack{\textbf{RMSE}\\\textbf{reduction (\%)}$\uparrow$}} &
{\shortstack{\textbf{95\% CI}\\\textbf{of reduction}}} &
{\shortstack{\textbf{Improved}\\\textbf{targets}}} \\
\midrule
Frozen backbone & \componentoff & \componentoff & 1.000 & 0.0 & -- & -- \\
+ GCU & \componenton & \componentoff & 0.956 & 4.4 & 2.6--6.6 & 86/129 \\
+ LVSE & \componentoff & \componenton & 0.893 & 10.7 & 6.9--14.8 & 78/129 \\
\rowcolor{RamanRow}
\textbf{RamanPFN} & \componenton & \componenton & \bestnum{0.868} & \bestnum{13.2} & \textbf{9.2--17.4} & \textbf{87/129} \\
\bottomrule
\end{tabular*}
\par\vspace{4pt}
\begin{minipage}{\textwidth}
\footnotesize\raggedright
Note: Checkmarks denote active components. The frozen TabPFN-3 backbone serves as the reference. RMSE is averaged over three repetitions within each target, and ratios to the reference are aggregated over 129 targets as geometric means with 95\% bootstrap confidence intervals. Arrows indicate the direction of better performance, and bold values mark the best result.
\end{minipage}
\end{widetable}
}
\wideblock{\begin{widefigure}
    \centering
    \includegraphics[width=\textwidth]{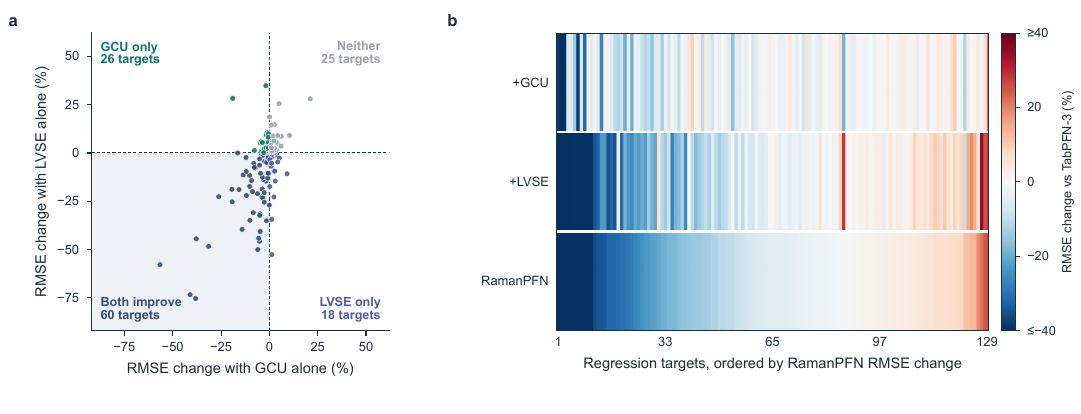}
    \caption{Ablation analysis of GCU and LVSE across 129 regression targets.
    (a)~Target-level RMSE changes for the GCU-augmented backbone and LVSE relative to the frozen TabPFN-3 backbone.
    (b)~Target-level RMSE changes for the GCU-augmented configuration, LVSE, and complete RamanPFN, ordered by complete RamanPFN and with color saturation at $\pm40\%$.}
    \label{fig:regression_ablation}
\end{widefigure}
}
\subsection{Evaluation metrics and statistical analysis}
\label{sec:evaluation_metrics}

Regression performance was measured by root-mean-square error (RMSE) and mean absolute error (MAE), multiclass classification by log loss, and binary classification by the area under the receiver operating characteristic curve (AUC). Each metric was first averaged across the three repetitions or outer folds within every task.

Relative error reductions of configuration $m$ against its reference $0$ across $T$ tasks were summarized by the geometric mean of task-level error ratios,
\begin{equation}
\label{eq:03-1}
\mathrm{Reduction}^{(m)}
=100\Bigl[1-\Bigl(\prod_{i=1}^{T}E_i^{(m)}\big/E_i^{(0)}\Bigr)^{1/T}\Bigr],
\end{equation}
where $E_i$ denotes the averaged RMSE or multiclass log loss on task $i$. A task was counted as improved when $E_i^{(m)}<E_i^{(0)}$. Confidence intervals were obtained from 20{,}000 bootstrap samples of the $T$ task-level log error ratios, using the 2.5th and 97.5th percentiles, and two-sided Wilcoxon signed-rank tests were applied to the same log ratios with Holm correction across contrasts within each analysis. For binary classification, the mean change in AUC is reported in percentage points together with task-level wins, ties, and losses.

Pairwise comparisons between RamanPFN and the 26 baselines under Protocol I used the symmetric RMSE advantage $100(\mathrm{RMSE}_b-\mathrm{RMSE}_a)/\max(\mathrm{RMSE}_a,\mathrm{RMSE}_b)$, where $a$ denotes RamanPFN and $b$ a comparison method. Target-level advantages were averaged within each of the 53 source datasets and tested with two-sided Wilcoxon signed-rank tests, using the Pratt treatment of zeros and Holm correction across the 26 prespecified contrasts.

Score and Elo were computed with the scoring procedure released with the RamanBench datasets, after averaging repetitions within each target. Score linearly maps the median negative RMSE on each target to zero and the best value to one, clips negative values to zero, and averages the result across the 129 targets. Elo ratings were computed from target-wise pairwise outcomes over 200 randomized target orderings with a $K$-factor of 32 and a logistic scale of 400, and were calibrated against the fixed reference point released with the datasets.

Channel coverage and co-occurrence were quantified with the balanced feature-subsampling routine of TabPFN-3 across the spectral widths of the 129 regression targets, with at most 500 features per forward pass and up to 32 estimators. For each target, 3{,}000 Monte Carlo trials sampled 128 random channel pairs and 128 random triplets. Marginal coverage recorded whether a channel appeared in at least one forward pass, and joint visibility recorded whether all members of a pair or triplet appeared together in at least one forward pass.

\section{Results}
\label{sec:results}

\wideblock{\begin{widefigure}
    \centering
    \includegraphics[width=\textwidth]{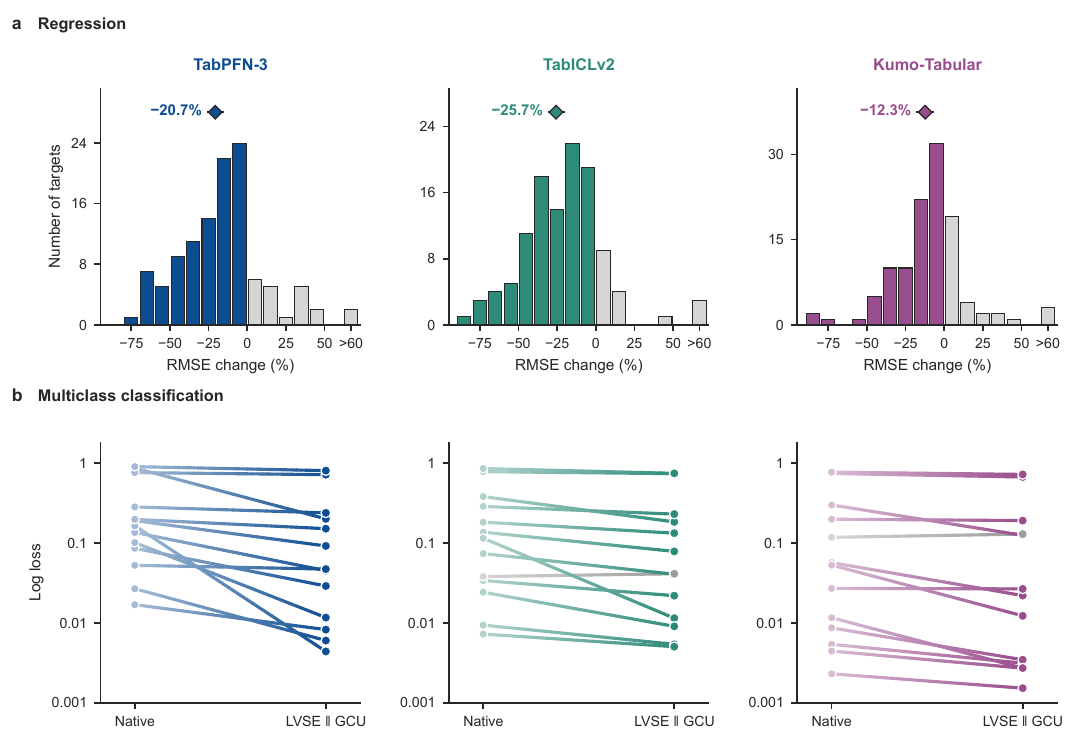}
    \caption{Generalization of RamanPFN across three tabular foundation models under Protocol II.
    (a)~Target-level RMSE change distributions relative to native inference across 114 regression targets, with gray bars denoting higher RMSE and diamonds and horizontal bars marking the geometric mean and 95\% bootstrap confidence interval.
    (b)~Task-level multiclass log loss on a logarithmic axis for native inference and RamanPFN, where gray lines denote higher log loss.}
    \label{fig:backbone_generalization}
\end{widefigure}
\par\vspace{8pt}\begin{widetable}
\centering
\rmfamily
\caption{Performance gains of RamanPFN across three tabular foundation models under Protocol II.}
\label{tab:backbone_generalization}
\footnotesize
\setlength{\tabcolsep}{2.0pt}
\renewcommand{\arraystretch}{1.20}
\begin{tabular*}{\textwidth}{@{\extracolsep{\fill}}l
  S[table-format=2.1] c S[table-format=2.1] c
  S[table-format=2.1] c c
  S[table-format=+1.2,retain-explicit-plus] c@{}}
\toprule
\multirow{2}{*}{\textbf{Backbone}} &
\multicolumn{4}{c}{\cellcolor{RegressionBand}\textbf{Regression (114 targets)}} &
\multicolumn{3}{c}{\cellcolor{ClassificationBand}\textbf{Multiclass (13 tasks)}} &
\multicolumn{2}{c}{\cellcolor{OverallBand}\textbf{Binary (8 tasks)}} \\
\cmidrule(lr){2-5}\cmidrule(lr){6-8}\cmidrule(lr){9-10}
& {\shortstack{\textbf{RMSE}\\\textbf{reduction (\%)}}} & {\textbf{95\% CI}} &
{\shortstack{\textbf{Median}\\\textbf{reduction (\%)}}} & {\textbf{Improved}} &
{\shortstack{\textbf{Log-loss}\\\textbf{reduction (\%)}}} & {\textbf{95\% CI}} & {\textbf{Improved}} &
{\shortstack{$\mathbf{\Delta}$\textbf{AUC}\\\textbf{(pp)}}} & {\textbf{W/T/L}} \\
\midrule
TabPFN-3 & 20.7 & 15.1--26.1 & 16.1 & 93/114 & 63.5 & 40.6--79.6 & 13/13 & +0.36 & 5/0/3 \\
TabICLv2 & 25.7 & 20.2--30.9 & 19.6 & 97/114 & 42.4 & 24.9--59.6 & 12/13 & +0.13 & 7/0/1 \\
Kumo-Tabular & 12.3 & 6.6--18.0 & 7.8 & 83/114 & 42.7 & 24.5--57.0 & 12/13 & +1.73 & 5/2/1 \\
\bottomrule
\end{tabular*}
\par\vspace{4pt}
\begin{minipage}{\textwidth}
\footnotesize\raggedright
Note: Each backbone with RamanPFN is compared with its native inference on raw spectra over three outer folds. Errors are averaged over folds within each task, and reductions are geometric means of error ratios with 95\% bootstrap confidence intervals. For binary tasks, $\Delta$AUC is the mean change in ROC AUC in percentage points, and W/T/L counts wins, ties, and losses.
\end{minipage}
\end{widetable}
}
\subsection{Coverage is not context}

\Cref{fig:spectral_mechanism}a reveals a pronounced divergence between marginal channel coverage and joint feature visibility across the 129 regression tasks as spectral dimensionality increases. On tasks with up to 594 channels, nearly all channel pairs appear together within at least one forward pass. At the median spectral width of 1{,}901 channels, marginal coverage remains complete, yet only 45.4\% of random channel pairs and 13.6\% of random triplets appear together in any forward pass. In the five widest tasks, which span 11{,}567 to 11{,}689 channels, joint visibility drops to 4.4\% for channel pairs and 0.19\% for triplets even though every individual channel is still sampled. Because the number of possible pairs and triplets grows quadratically and cubically with spectral width, feature subsampling increasingly separates covarying peaks across different forward passes.

The support maps in \Cref{fig:spectral_mechanism}b compare this subsampling behavior with the coordinates constructed by GCU and LVSE on a 961-channel gasoline spectrum. Eight TabPFN-3 forward passes observe interleaved, fragmented channel subsets across the wavenumber axis. By contrast, the six highest-variance GCU components each span the entire spectrum while concentrating their strongest weights on distinct sets of covarying bands, and LVSE partitions the axis into 61 contiguous intervals that keep local band profiles intact, including the highlighted C--H stretching interval from 2{,}937 to 2{,}995~cm$^{-1}$.

\subsection{Physical interpretability of GCU and LVSE}

To examine whether these coordinates capture chemical structure, we inspected the learned GCU components and LVSE modes on two mixture systems. In \Cref{fig:physical_interpretation}a, non-negative combinations of the unsupervised GCU basis spectra reconstruct the pure reference spectra of sucrose, maltose, glucose, and fructose from the low signal-to-noise sugar mixture dataset~\citep{georgiev2024} with relative errors between 0.26\% and 0.40\%. Across the mixture series, the corresponding GCU coordinates track the volume fractions of the individual sugars, with Pearson correlation coefficients of 0.94 for sucrose, 0.99 for maltose, 0.79 for glucose, and 0.90 for fructose.

A similar correspondence emerges in multicomponent gasoline spectra without using concentration labels during factorization. As shown in \Cref{fig:physical_interpretation}b, individual GCU components isolate spectral signatures dominated by distinct bands at 885~cm$^{-1}$ for ethanol, 727~cm$^{-1}$ for methyl \textit{tert}-butyl ether (MTBE), and 742~cm$^{-1}$ for ethyl \textit{tert}-butyl ether (ETBE). The resulting sample coordinates rise with the measured oxygenate contents, with correlation coefficients of 0.67 for ethanol, 0.93 for MTBE, and 0.90 for ETBE. All four admissible unmixings recover the same ETBE component profile, with pairwise cosine similarities of 0.91 to 0.97.

Complementing full-spectrum unmixing, \Cref{fig:physical_interpretation}c illustrates how LVSE resolves fine-grained band morphology within the 16-channel C--H stretching interval spanning 2{,}937 to 2{,}995~cm$^{-1}$. The first three orthogonal modes account for 71.8\%, 25.3\%, and 2.3\% of the local variance, capturing 99.3\% in total. The leading mode follows the overall band envelope and shows only a weak association with ETBE content, with an absolute correlation of 0.15, whereas the second mode captures asymmetric contrast across the band and reaches an absolute correlation of 0.86. The third mode resolves subtler shoulder structure. Retaining multiple local modes per interval thus preserves within-band shape variations that a single dominant direction would discard.

\subsection{Regression performance against comparison methods}

Under Protocol I, RamanPFN outperformed all 26 comparison methods across every aggregate regression measure in \Cref{tab:regression_comparison}. Across the 129 targets, RamanPFN achieved the lowest mean RMSE of 9.171, the lowest mean MAE of 6.910, the smallest mean gap to the best method on each target at 7.2\%, and the best average rank of 2.65, while leading in both Score at 0.8694 and Elo rating at 1{,}702. Among the comparison methods, TabPFN~v2 achieved the second-best mean RMSE of 9.574, average rank of 5.29, and Elo of 1{,}489, while TabPFN~v2.5 achieved the second-best mean MAE of 7.227, gap to best of 28.2\%, and Score of 0.6316.

Relative to the frozen TabPFN-3 backbone, RamanPFN reduced RMSE by a geometric mean of 13.2\%, with a 95\% bootstrap confidence interval of 9.2\% to 17.4\%, and lowered prediction error on 87 of the 129 targets, as shown in \Cref{fig:regression_performance}a. In the target-wise comparison of \Cref{fig:regression_performance}b, where the strongest baseline was selected separately for every target from the 26 comparison methods, RamanPFN still achieved a lower RMSE on 77 of the 129 targets.

This advantage held when aggregated over the 53 source datasets in \Cref{fig:regression_performance}c. Giving each dataset equal weight, RamanPFN achieved a positive mean symmetric RMSE advantage over every comparison method, ranging from 19.5\% over TabPFN~v2.5 and 22.5\% over TabPFN~v2 to 54.0\% over RamanTransformer. All 26 prespecified contrasts were statistically significant after Holm correction, with adjusted $P\le4.16\times10^{-7}$.

\subsection{Contributions of global and local spectral learning}

\Cref{tab:regression_ablation} summarizes the ablation results under Protocol I across the 129 regression targets. Relative to the frozen backbone, augmenting the backbone with GCU coordinates reduced RMSE by a geometric mean of 4.4\%, with a 95\% confidence interval of 2.6\% to 6.6\%, improving 86 of the 129 targets. Using LVSE coordinates reduced RMSE by 10.7\%, with a 95\% confidence interval of 6.9\% to 14.8\%, improving 78 targets. Combining both scales in complete RamanPFN reduced RMSE by 13.2\% with 87 improved targets, an additional 2.75\% reduction beyond LVSE alone at $P=3.1\times10^{-6}$.

The target-level comparison in \Cref{fig:regression_ablation}a shows why combining the two spectral scales outperforms either module alone. Of the 129 targets, 60 improved under both the GCU-augmented and LVSE configurations, whereas 26 improved only with GCU, 18 only with LVSE, and 25 with neither. Complete RamanPFN improved 59 of the 60 targets that benefited from both modules and 85 of the 104 targets that benefited from at least one, and the ordered target profiles in \Cref{fig:regression_ablation}b show how full-spectrum compositional coordinates and local vibrational coordinates combine across the regression tasks.

\subsection{Generalization across backbones and tasks}

\Cref{tab:backbone_generalization} and \Cref{fig:backbone_generalization} report the Protocol II evaluation, in which RamanPFN was applied with the same fixed configuration to TabPFN-3, TabICLv2, and Kumo-Tabular. On the 114 regression targets shown in \Cref{fig:backbone_generalization}a, RamanPFN reduced RMSE by a geometric mean of 20.7\% on TabPFN-3, with a 95\% confidence interval of 15.1\% to 26.1\%, by 25.7\% on TabICLv2, with an interval of 20.2\% to 30.9\%, and by 12.3\% on Kumo-Tabular, with an interval of 6.6\% to 18.0\%. Median target-level reductions reached 16.1\%, 19.6\%, and 7.8\%, with lower RMSE on 93, 97, and 83 of the 114 targets, respectively.

Larger gains appeared in multiclass classification, where RamanPFN reduced log loss by 63.5\% on TabPFN-3, 42.4\% on TabICLv2, and 42.7\% on Kumo-Tabular, and the task-level trajectories in \Cref{fig:backbone_generalization}b show this decrease on 13, 12, and 12 tasks for the three backbones. All six regression and multiclass confidence intervals excluded zero, and the reductions remained significant after Holm correction. On the binary classification tasks, RamanPFN increased the mean ROC AUC by 0.36, 0.13, and 1.73 percentage points.

\section{Discussion}
\label{sec:discussion}

The results show that the difficulty of applying tabular foundation models to Raman spectra lies in how spectral evidence reaches the model. Feature subsampling covers every channel but places distant covarying bands in different forward passes and splits contiguous band profiles, and this separation grows with spectral width. GCU and LVSE reduce spectra of up to 11{,}689 channels to at most 224 coordinates, all of which the backbone observes within every forward pass. Each GCU coordinate collects covarying responses from distant bands into one abundance-like variable, following the additive-mixture reasoning of constrained Raman unmixing~\citep{georgiev2024}. Each LVSE coordinate describes one direction of variation within a contiguous interval, and retaining several modes per interval extends locality-preserving spectral encoders and single-direction segment compression~\citep{hypersigma2025,gotabpfn2026}. Physical structure thus determines which channels are combined, and the backbone learns from the combined evidence with its parameters unchanged.

The ablation separates the contributions of the two scales. Augmenting the backbone with GCU coordinates lowered RMSE on more targets than LVSE coordinates alone, whereas LVSE produced the larger average reduction. Global coordinates therefore provide moderate gains across many targets, and local coordinates provide larger gains on a narrower set. Since 44 targets improved under only one of the two configurations, neither scale substitutes for the other, and their combination produced the largest reduction relative to the frozen backbone.

Beyond accuracy, the coordinates remain open to chemical inspection. GCU components learned without concentration labels reconstructed the reference spectra of four sugars and isolated the dominant bands of three gasoline oxygenates, and the corresponding coordinates followed the measured contents. Each global coordinate can therefore be read as an approximate component abundance, with a basis spectrum that an analyst can compare against reference spectra. Unlabeled spectra determine neither the number of components nor a unique non-negative intensity scale, and averaging predictions over the four admissible unmixings removes the need to fix either choice for each task. These unmixings recovered nearly identical ETBE components, so the averaging preserves the chemical reading of the coordinates. LVSE modes distinguish the band envelope from asymmetric changes within a band, and in the C--H stretching interval the ETBE signal appeared in the second mode. These properties tie the inputs of the backbone to identifiable components and spectral intervals, allowing analysts to check model outputs against chemical knowledge in quality control and diagnostic applications.

The same fixed configuration lowered regression error and multiclass log loss for three tabular foundation models developed independently by different groups. The gains therefore come from the physical encoding of the spectra and carry over across backbones. Related studies have adapted tabular foundation models to scientific data through multi-fidelity information and composition or structure descriptors~\citep{mftabpfn2026,iclfm2026}. In RamanPFN, domain knowledge enters through the spectral axis of each measurement, while the pretrained model supplies a learned inference procedure~\citep{hollmann2025,tabicl2025}. Because the backbone parameters remain fixed, RamanPFN can adopt newer tabular foundation models as they are released without fine-tuning.

This study focused on Raman spectroscopy, and the two properties that RamanPFN encodes also characterize other ordered spectral measurements. Near-infrared and mid-infrared absorbance spectra of mixtures follow additive mixing under the Beer--Lambert law, and tabular foundation models have already been applied to near-infrared calibration~\citep{reiter2026nir}. Future work will extend RamanPFN to these modalities and to hyperspectral imaging, where linear unmixing of pixel spectra is well established~\citep{rasti2024hysupp} and spatial structure accompanies each spectrum.

\section{Conclusion}
\label{sec:conclusion}

This study examined how tabular foundation models process high-dimensional Raman spectra and showed that feature subsampling preserves channel coverage while joint channel visibility collapses as spectral width grows. RamanPFN addresses this limitation by combining full-spectrum compositional coordinates from GCU with local vibrational coordinates from LVSE and by marginalizing predictions over admissible unmixings while the pretrained backbone remains fixed. Evaluated on 74 public Raman datasets comprising more than 330{,}000 spectra from four application domains, RamanPFN achieved the lowest aggregate RMSE and MAE and the best Score and Elo against 26 comparison methods across 129 regression targets. It reduced RMSE by 13.2\% relative to its frozen backbone, and its dataset-level gains over all 26 methods remained significant after Holm correction. Applied with the same configuration to TabPFN-3, TabICLv2, and Kumo-Tabular, it reduced RMSE by 12.3\%--25.7\% across 114 regression targets and multiclass log loss by 42.4\%--63.5\%. GCU components reproduced the reference spectra of mixture constituents and followed measured contents without concentration labels, and LVSE modes resolved composition-dependent changes in band shape, so the coordinates passed to the backbone can be interpreted chemically. RamanPFN thus provides an accurate and interpretable approach to Raman-based quantitative analysis and sample identification that can directly benefit from future tabular foundation models.

\bibliographystyle{elsarticle-num-names}
\bibliography{references}
\end{document}